\documentclass{article}

\PassOptionsToPackage{numbers, compress}{natbib}
\usepackage[final]{nips_2017}

\usepackage{color}
\usepackage[utf8]{inputenc} % allow utf-8 input
\usepackage[T1]{fontenc}    % use 8-bit T1 fonts
\usepackage{hyperref}       % hyperlinks
\usepackage{url}            % simple URL typesetting
\usepackage{booktabs}       % professional-quality tables
\usepackage{amsfonts}       % blackboard math symbols
\usepackage{nicefrac}       % compact symbols for 1/2, etc.
\usepackage{microtype}      % microtypography
\usepackage{mmstyle}
\usepackage{caption}
\usepackage{multirow}
\usepackage{amssymb}
\usepackage{amsmath}
\usepackage{graphicx}

\title{Contrastive Learning for Image Captioning}

\author{Bo Dai\qquad\qquad\qquad\qquad~Dahua Lin \\
Department of Information Engineering, The Chinese University of Hong Kong\\
{\footnotesize\texttt{db014@ie.cuhk.edu.hk}\qquad\texttt{dhlin@ie.cuhk.edu.hk}}\\
}

\begin{document}
% \nipsfinalcopy is no longer used

\maketitle

% !TEX root = main.tex

\begin{abstract}
Image captioning, a popular topic in computer vision,
has achieved substantial progress in recent years.
However, the \emph{distinctiveness} of natural descriptions
is often overlooked in previous work.
It is closely related to the quality of captions,
as distinctive captions are more likely to describe images with their unique aspects.
In this work, we propose a new learning method,
Contrastive Learning (CL),
for image captioning.
Specifically,
via two constraints formulated on top of a reference model,
the proposed method can encourage distinctiveness,
while maintaining the overall quality of the generated captions.
We tested our method on two challenging datasets,
where it improves the baseline model
by significant margins.
We also showed in our studies that
the proposed method is generic and can be used for models with various structures.
\end{abstract}

% !TEX root = main.tex

\section{Introduction}
\label{sec:intro}
%background
Image captioning, a task to generate natural descriptions of images, has been
an active research topic in computer vision and machine learning.
Thanks to the advances in deep neural networks, especially the wide adoption of RNN and LSTM,
there has been substantial progress on this topic in
recent years~\cite{vinyals2015show, xu2015show, lu2016knowing, rennie2016self}.
%
%motivation (limitation)
However, studies~\cite{dai2017towards,fang2015captions,devlin2015exploring,kuznetsova2014treetalk}
have shown that even the captions generated by state-of-the-art models still leave a lot to be desired.
Compared to human descriptions,
machine-generated captions are often quite rigid and tend to favor
a \emph{``safe''} (\ie~matching parts of the training captions in a word-by-word manner)
but \emph{restrictive} way.
As a consequence, captions generated for different images,
especially those that contain objects of the same categories,
are sometimes very similar~\cite{dai2017towards}, despite their differences in other aspects.

% Distinctiveness
%
We argue that \textbf{distinctiveness}, a property often overlooked in previous work,
is significant in natural language descriptions.
To be more specific, when people describe an image, they often mention or even emphasize
the \emph{distinctive} aspects of an image that distinguish it from others.
With a distinctive description, someone can easily identify the image it is referring to, among
a number of similar images.
In this work, we performed a \emph{self-retrieval}
study (see Section~\ref{sec:selfretrieval}), which reveals the lack of distinctiveness
affects the quality of descriptions.

% Such resemblance deviates machine-generated captions from an important property of natural image-caption pairs,
% \textbf{distinctiveness},
% which indicates the one-to-one relationships between captions and images.
% Specifically, in natural scenarios,
% an image can often be easily retrieved when its caption is presented,
% as the caption contains \emph{key and distinct} semantics of the image.
% Different images should contain different distinct semantics, resulting in the diversity of captions.
% Therefore, high resemlance in captions is a strong indication
% that they only capture the \emph{common} semantics,
% which should be avoided.
% Intuitively, the most common semantics across all images is "something in this image",
% which is trivially right to be a caption,
% but is discouraged in human's mind.

% Technical reasons
%
From a technical standpoint, the lack of \emph{distinctiveness} is partly related
to the way that the captioning model was learned. A majority of image captioning
models are learned by Maximum Likelihood Estimation (MLE),
where the probabilities of training captions conditioned on corresponding
images are maximized.
While well grounded in statistics, this approach does not explicitly promote
distinctiveness. Specifically, the differences among the captions of different images
are not explicitly taken into account.
We found empirically that the resultant captions highly resemble the training set
in a word-by-word manner, but are not \emph{distinctive}.

% However, most image captioning models are trained by maximum likelihood estimation (MLE),
% where the probabilities $p(c|I)$ of ground-truth image-caption pairs are maximized.
% And the probabilities $p(c_/|I)$ of negative image-caption pairs,
% \eg an image with a caption that describes another different image,
% are left out.
% The omission of such discriminative cues may lead to insufficient estimation of
% the distinctiveness of ground-truth image-caption pairs,
% which we found empirically downgrades the performance of image captioning models.

% introduce our proposal first, then discuss the high-level limitation of similar works
In this paper, we propose \textbf{Contrastive Learning (CL)},
a new learning method for image captioning, which explicitly encourages
\emph{distinctiveness}, while maintaining the overall quality of the
generated captions.
Specifically, it employs a baseline, \eg~a state-of-the-art model, as a \emph{reference}.
During learning, in addition to true image-caption pairs, denoted as $(I, c)$, this method
also takes as input \emph{mismatched pairs}, denoted as $(I, c_/)$, where $c_/$ is a caption
describing another image.
Then, the target model is learned to meet two goals, namely
(1) giving higher probabilities $p(c | I)$ to positive pairs,
and (2) lower probabilities $p(c_/ | I)$ to negative pairs, compared to the reference model.
The former ensures that the overall performance of the target model is not inferior to the reference;
while the latter encourages distinctiveness.

It is noteworthy that the proposed learning method (CL) is generic.
While in this paper, we focused on models based on
recurrent neural networks~\cite{vinyals2015show, lu2016knowing},
the proposed method can also generalize well to models based on
other formulations, \eg~probabilistic graphical models~\cite{farhadi2010every, kulkarni2013babytalk}.
Also, by choosing the state-of-the-art model as the reference model in CL,
one can build on top of the latest advancement in image captioning to obtain improved performances.

% !TEX root = main.tex

\section{Related Work}
\label{sec:relwork}

%captioning models
  %old way
  %lstms
  %attention
  %attention with results from other task (e.g. attribute recognition)
%learning methods

\paragraph{Models for Image Captioning}
The history of image captioning can date back to decades ago.
Early attempts are mostly based on detections,
which first detect visual concepts (\eg objects and their attributes) \cite{kulkarni2013babytalk, farhadi2010every}
followed by template filling \cite{kulkarni2013babytalk}~or nearest neighbor retrieving for caption generation \cite{devlin2015exploring, farhadi2010every}.
With the development of neural networks,
a more powerful paradigm, \emph{encoder-and-decoder}, was proposed by \cite{vinyals2015show},
which then becomes the core of most state-of-the-art image captioning models.
It uses a CNN \cite{simonyan2014very}~to represent the input image with a feature vector,
and applies a LSTM net~\cite{hochreiter1997long} upon the feature to generate words one by one.

Based on the encoder-and-decoder, many variants are proposed,
where \emph{attention mechanism} \cite{xu2015show}~appears to be the most effective add-on.
Specifically, attention mechanism replaces the feature vector with a set of feature vectors,
such as the features from different regions \cite{xu2015show}~, and those under different conditions \cite{zhou2016image}.
It also uses the LSTM net to generate words one by one,
where the difference is that
at each step, a mixed guiding feature over the whole feature set,
will be \emph{dynamically} computed.
In recent years, there are also approaches combining attention mechanism and detection.
Instead of doing attention on features,
they consider the attention on a set of detected visual concepts,
such as attributes \cite{yao2016boosting} and objects \cite{you2016image}.

Despite of the specific structure of any image captioning model,
it is able to give $p(c|I)$, the probability of a caption conditioned on an image.
Therefore, all image captioning models can be used as the target or the reference in CL method.

\paragraph{Learning Methods for Image Captioning}
Many state-of-the-art image captioning models adopt \emph{Maximum Likelihood Estimation (MLE)} as their learning method,
which maximizes the conditional log-likelihood of the training samples, as:
\begin{align}
	\sum_{(c_i,I_i) \in \cD} \sum^{T_i}_{t = 1} \ln p(w^{(t)}_i|I_i, w^{(t - 1)}_i, ..., w^{(1)}_i, \vtheta),
\end{align}
where $\vtheta$ is the parameter vector, $I_i$ and $c_i = (w^{(1)}_i, w^{(2)}_i, ..., w^{(T_i)}_i)$ are a training image and its caption.
Although effective, some issues, including high resemblance in model-gerenated captions, are observed \cite{dai2017towards}~on models learned by MLE.

Facing these issues, alternative learning methods are proposed in recent years.
Techniques of reinforcement learning (RL) have been introduced in image captioning by \cite{rennie2016self}~and \cite{liu2016optimization}.
RL sees the procedure of caption generation as a procedure of sequentially sampling actions (words) in a policy space (vocabulary).
The rewards in RL are defined to be evaluation scores of sampled captions.
Note that distinctiveness has not been considered in both approaches, RL and MLE.

Prior to this work,
some relevant ideas have been explored \cite{vedantam2017context,mao2016generation,dai2017towards}.
Specifically, \cite{vedantam2017context,mao2016generation}~proposed an introspective learning (IL) approach
that learns the target model by comparing its outputs on $(I,c)$ and $(I_/,c)$.
Note that IL uses the target model itself as a reference.
On the contrary, the reference model in CL provides more \emph{independent} and \emph{stable} indications about distinctiveness.
In addition, $(I_/, c)$ in IL is pre-defined and fixed across the learning procedure,
while the negative sample in CL, \ie $(I,c_/)$, is \emph{dynamically} sampled, making it more diverse and random.
Recently, Generative Adversarial Networks (GAN) was also adopted for image captioning \cite{dai2017towards},
which involves an evaluator that may help promote the distinctiveness.
However, this evaluator is \emph{learned} to \emph{directly} measure the distinctiveness as a parameterized approximation,
and the approximation accuracy is not ensured in GAN.
In CL, the \emph{fixed} reference provides stable \emph{bounds} about the distinctiveness,
and the bounds are supported by the model's performance on image captioning.
Besides that, \cite{dai2017towards} is specifically designed for models that generate captions word-by-word,
while CL is more generic.

% !TEX root = main.tex

\section{Background}
\label{sec:bg}

Our formulation is partly inspired by
\emph{Noise Contrastive Estimation (NCE)}~\cite{gutmann2012noise}.
NCE is originally introduced for estimating probability distributions,
where the partition functions can be difficult or even infeasible to compute.
To estimate a parametric distribution $p_m(.;\vtheta)$,
which we refer to as the \emph{target} distribution,
NCE employs not only the observed samples $X = (\vx_1, \vx_2, ..., \vx_{T_m})$,
but also the samples drawn from a \emph{reference} distribution $p_n$,
denoted as $Y = (\vy_1, \vy_2, ..., \vy_{T_n})$.
Instead of estimating $p_m(.;\vtheta)$ directly,
NCE estimates the density ratio $p_m / p_n$
by training a classifier based on logistic regression.

Specifically, let $U = (\vu_1, ..., \vu_{T_m + T_n})$ be the union of $X$ and $Y$.
A binary class label $C_t$ is assigned to each $u_t$,
where $C_t = 1$ if $u_t \in X$ and $C_t = 0$ if $u_t \in Y$.
The posterior probabilities for the class labels are therefore
\begin{align}
	P(C = 1|\vu,\vtheta) = \frac{p_m(\vu;\vtheta)}{p_m(\vu;\vtheta) + \nu p_n(\vu)},  \qquad P(C = 0|\vu,\vtheta) = \frac{\nu p_n(\vu)}{p_m(\vu;\vtheta) + \nu p_n(\vu)},
\end{align}
where $\nu = T_n / T_m$.
Let $G(\vu;\vtheta) = \ln p_m(\vu;\vtheta) - \ln p_n(\vu)$ and
$h(\vu,\vtheta) = P(C = 1|\vu,\vtheta)$, then we can write
\begin{equation}
	h(\vu;\vtheta) = r_\nu(G(\vu;\vtheta)), \quad
	\text{ with } \quad
	r_\nu(z) = \frac{1}{1 + \nu\exp(-z)}. \label{eq:logistic}
\end{equation}
The objective function of NCE is the joint conditional log-probabilities
of $C_t$ given the samples $U$, which can be written as
\begin{equation}
	\cL(\vtheta;X,Y)
	=
	\sum^{T_m}_{t = 1} \ln[h(\vx_t;\vtheta)] + \sum^{T_n}_{t = 1} \ln[1 - h(\vy_t;\vtheta)].
\end{equation}
Maximizing this objective with respect to $\vtheta$ leads to an estimation
of $G(\cdot;\vtheta)$, the logarithm of the density ratio $p_m / p_n$.
As $p_n$ is a known distribution, $p_m(:|\vtheta)$ can be readily derived.

% The class labels $C_t$ are assumed Bernoulli distributed and independent in NCE,
% and the conditional log-likelihood is thus given by
% \begin{align}
% 	\cL(\vtheta;X,Y) & = \sum^{T_m + T_n}_{t = 1} C_t \ln P(C_t = 1|\vu,\vtheta) + (1 - C_t) \ln P(C_t = 0 | \vu,\vtheta) \notag \\
% 			& = \sum^{T_m}_{t = 1} \ln[h(\vx_t;\vtheta)] + \sum^{T_n}_{t = 1} \ln[1 - h(\vy_t;\vtheta)] .
% \end{align}

% !TEX root = main.tex

\section{Contrastive Learning for Image Captioning}
\label{sec:frmwork}

\begin{figure}
	\centering
	\includegraphics[width=\textwidth]{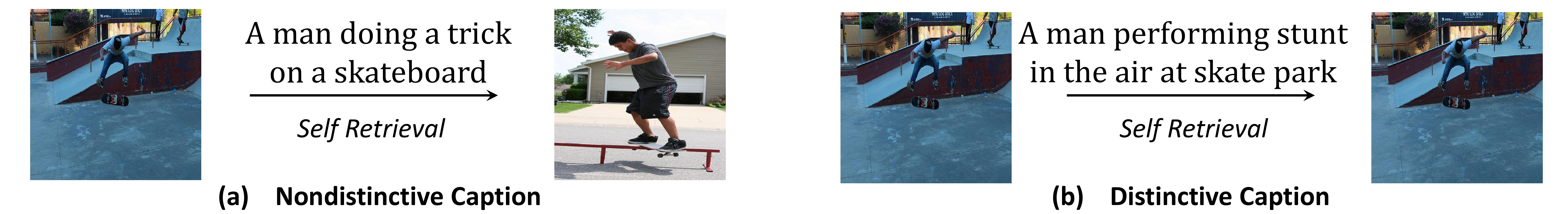}
	\caption{\small This figure illustrates respectively a nondistinctive and distinctive captions of an image,
                 where the nondistinctive one fails to retrieve back the original image in \emph{self retrieval} task.}
	\label{fig:selfretrieval}
	\vspace{-1mm}
\end{figure}

\begin{table}
\small
\centering
\begin{tabular}{lcccccc}
\toprule
	& \multicolumn{4}{c}{Self Retrieval Top-K Recall} & \multicolumn{2}{c}{Captioning} \\
\cmidrule(l{5pt}r{5pt}){2-5} \cmidrule(l{5pt}r{5pt}){6-7} 
Method               & 1 & 5 & 50 & 500 & ROUGE\_L & CIDEr \\ 
\midrule
Neuraltalk2 \cite{karpathy2015deep} & 0.02 & 0.32 & 3.02 & 27.50 & 0.652 & 0.827 \\
AdaptiveAttention \cite{lu2016knowing} & 0.10 & 0.96 & 11.76 & 78.46 & 0.689 & 1.004 \\
AdaptiveAttention + CL & 0.32 & 1.18 & 11.84 & 80.96 & 0.695 & 1.029 \\
\bottomrule
\end{tabular}
\vspace{1mm}
\caption{\small This table lists results of self retrieval and captioning of different models. 
The results are reported on standard MSCOCO test set.
See sec \ref{sec:selfretrieval} for more details.}
\label{tab:selfretrieval}
\vspace{-5mm}
\end{table}

Learning a model by characterizing desired properties relative to a strong baseline
is a convenient and often quite effective way in situations
where it is hard to describe these properties directly.
Specifically, in image captioning,
it is difficult to characterize
the distinctiveness of natural image descriptions
via a set of rules,
without running into the risk that some subtle but significant points are missed.
Our idea in this work is to introduce a baseline model as a reference,
and try to enhance the distinctiveness on top, while maintaining the
overall quality of the generated captions.

In the following we will first present an empirical study
on the correlation between \emph{distinctiveness} of its generated captions and the
\emph{overall performance} of a captioning model.
Subsequently, we introduce the main framework of \emph{Contrastive Learning}
in detail.

\subsection{Empirical Study: Self Retrieval}
\label{sec:selfretrieval}

In most of the existing learning methods of image captioning,
models are asked to generate a caption that best describes the semantics of a given image.
In the meantime, \textbf{distinctiveness} of the caption,
which, on the other hand, requires the image to be the best matching \emph{among all images}
for the caption, has not been explored.
However, distinctiveness is crucial for high-quality captions.
A study by Jas \cite{jas2015image} showed that \emph{specificity} is common in human descriptions, 
which implies that image descriptions often involve distinctive aspects.
Intuitively, a caption satisfying this property is very likely to contain key and unique content of the image,
so that the original image could easily be retrieved when the caption is presented.

To verify this intuition, we conducted an empirical study which we refer to as \emph{self retrieval}.
In this experiment, we try to retrieve the original image given its model-generated caption
and investigate top-$k$ recalls, as illustrated in Figure \ref{fig:selfretrieval}.
Specifically, we randomly sampled $5,000$ images $(I_1, I_2, ..., I_{5000})$ from standard MSCOCO \cite{lin2014microsoft}~test set
as the experiment benchmark.
For an image captioning model $p_m(:,\vtheta)$,
we first ran it on the benchmark to get corresponding captions $(c_1, c_2, ..., c_{5000})$ for the images.
After that,
using each caption $c_t$ as a query,
we computed the conditional probabilities $(p_m(c_t|I_1), p_m(c_t|I_2), ..., p_m(c_t|I_{5000}))$,
which were used to get a ranked list of images, denoted by $\vr_t$.
Based on all ranked lists, we can compute top-$k$ recalls,
which is the fraction of images within top-$k$ positions of their corresponding ranked lists.
The top-$k$ recalls are good indicators of how well a model captures
the distinctiveness of descriptions.

In this experiment, we compared three different models, including 
\emph{Neuraltalk2} \cite{karpathy2015deep} and \emph{AdaptiveAttention} \cite{lu2016knowing} that are learned by MLE,
as well as \emph{AdaptiveAttention} learned by our method.
The top-$k$ recalls are listed in Table~\ref{tab:selfretrieval},
along with overall performances of these models in terms of
\emph{Rouge}~\cite{lin2004rouge}~and \emph{Cider}~\cite{vedantam2015cider}.
These results clearly show that
the recalls of self retrieval are positively correlated to the performances of image captioning models
in classical captioning metrics.
Although most of the models are not explicitly learned to promote distinctiveness,
the one with better recalls of self retrieval,
which means the generated-captions are more distinctive,
performs better in the image captioning evaluation.
Such positive correlation clearly demonstrates the significance of
\emph{distinctiveness} to captioning performance.

\subsection{Contrastive Learning}

In Contrastive Learning (CL),
we learn a target image captioning model $p_m(:;\vtheta)$ with parameter $\vtheta$
by constraining its behaviors relative to a reference model $p_n(:;\vphi)$ with parameter $\vphi$.
The learning procedure requires two sets of data:
(1) the observed data $X$,
which is a set of ground-truth image-caption pairs $((c_1, I_1), (c_2, I_2), ..., (c_{T_m}, I_{T_m}))$,
and is readily available in any image captioning dataset,
(2) the noise set $Y$,
which contains mismatched pairs $((c_{/1}, I_1), (c_{/2}, I_2), ..., (c_{/T_n}, I_{T_n}))$,
and can be generated by randomly sampling $c_{/t} \in \cC_{/I_t}$ for each image $I_t$,
where $\cC_{/I_t}$ is the set of all ground-truth captions except captions of image $I_t$.
We refer to $X$ as \emph{positive pairs} while $Y$ as \emph{negative pairs}.

For any pair $(c,I)$,
the target model and the reference model will respectively give
their estimated conditional probabilities $p_m(c|I,\vtheta)$ and $p_n(c|I,\vphi)$.
We wish that $p_m(c_t|I_t, \vtheta)$ is greater than $p_n(c_t|I_t, \vphi)$ for any positive pair $(c_t, I_t)$,
and vice versa for any negative pair $(c_{/t}, I_t)$.
Following this intuition, our initial attempt was to
define $D((c,I);\vtheta, \vphi)$, the difference between $p_m(c|I, \vtheta)$ and $p_n(c|I, \vphi)$, as
\begin{align}
	D((c,I);\vtheta, \vphi) = p_m(c|I, \vtheta) - p_n(c|I, \vphi),
\end{align}
and set the loss function to be:
\begin{align}
	\cL^\prime(\vtheta;X, Y, \vphi) =
	\sum^{T_m}_{t = 1} D((c_t, I_t);\vtheta, \vphi)
	- \sum^{T_n}_{t = 1} D((c_{/t}, I_t);\vtheta, \vphi)
	\label{eq:dsum}.
\end{align}
In practice, this formulation would meet with several difficulties.
First,
$p_m(c|I, \vtheta)$ and $p_n(c|I, \vphi)$ are very small ($\sim 1e$-$8$),
which may result in numerical problems.
Second,
Eq~\eqref{eq:dsum} treats easy samples,
hard samples, and mistaken samples equally.
This, however, is not the most effective way.
For example, when $D((c_t, I_t);\vtheta, \vphi) \gg 0$ for some positive pair,
further increasing $D((c_t, I_t);\vtheta, \vphi)$
is probably not as effective as updating $D((c_{t^\prime}, I_{t^\prime});\vtheta,\vphi)$
for another positive pair,
for which $D((c_{t^\prime}, I_{t^\prime});\vtheta,\vphi)$ is much smaller.

To resolve these issues, we adopted an alternative formulation
inspired by NCE (sec \ref{sec:bg}), where
we replace the difference function $D((c, I);\vtheta, \vphi)$
with a log-ratio function $G((c,I);\vtheta, \vphi)$:
\begin{align}
	G((c,I);\vtheta, \vphi) = \ln p_m(c|I, \vtheta) - \ln p_n(c|I, \vphi),
\end{align}

and further use a logistic function $r_\nu$ (Eq(\ref{eq:logistic})) after $G((c,I);\vtheta, \vphi)$ to saturate the influence of easy samples.
Following the notations in NCE, we let $\nu = T_n / T_m$,
and turn $D((c, I);\vtheta, \vphi)$ into:
\begin{align}
	h((c,I);\vtheta, \vphi) = r_\nu(G((c,I);\vtheta, \vphi))).
\end{align}

Note that $h((c,I);\vtheta, \vphi) \in (0, 1)$. Then, we define our updated loss function as:
\begin{align}
	\cL(\vtheta;X,Y,\vphi) = \sum^{T_m}_{t = 1} \ln[h((c_t,I_t);\vtheta,\vphi)] + \sum^{T_n}_{t = 1}\ln[1 - h((c_{/t},I_t); \vtheta, \vphi)] \label{eq:hsum}.
\end{align}

For the setting of $\nu = T_n / T_m$,
we choose $\nu = 1$, \ie~$T_n = T_m$, to ensure balanced influences
from both positive and negative pairs.
This setting consistently yields good performance in our experiments.
Furthermore, we copy $X$ for $K$ times and sample $K$ different $Y$s,
in order to involve more diverse negative pairs without overfitted to them.
In practice we found $K = 5$ is sufficient to make the learning stable.
Finally, our objective function is defined to be
\begin{align}
	J(\vtheta) = \frac{1}{K} \frac{1}{T_m} \sum^K_{k = 1} \cL(\vtheta;X,Y_k,\vphi).
\end{align}
%analysis of the terms
Note that $J(\vtheta)$ attains its upper bound $0$ if positive and negative pairs can be
perfectly distinguished, namely,
for all $t$, $h((c_t,I_t); \vtheta, \vphi) = 1$ and $h((c_{/t}, I_t); \vtheta, \vphi) = 0$.
In this case, $G((c_t, I_t);\vtheta, \vphi) \to \infty$ and $G((c_{/t},I_t); \vtheta, \vphi) \to -\infty$,
which indicates the target model will give higher probability $p(c_t|I_t)$ and lower probability $p(c_{/t}|I_t)$,
compared to the reference model.
Towards this goal, the learning process would encourage \emph{distinctiveness} by suppressing
negative pairs, while maintaining the overall performance by maximizing the probability values
on positive pairs.

\subsection{Discussion}
\label{sec:discuss}
%MLE,NCE,GAN
Maximum Likelihood Estimation (MLE) is a popular learning method in the area of image captioning \cite{vinyals2015show, xu2015show, lu2016knowing}.
The objective of MLE is to maximize \emph{only} the probabilities of ground-truth image-caption pairs,
which may lead to some issues \cite{dai2017towards},
including high resemblance in generated captions.
While in CL, the probabilities of ground-truth pairs are \emph{indirectly} ensured by the positive constraint
(the first term in Eq(\ref{eq:hsum})),
and the negative constraint (the second term in Eq(\ref{eq:hsum}))
suppresses the probabilities of mismatched pairs,
forcing the target model to also learn from distinctiveness.

Generative Adversarial Network (GAN) \cite{dai2017towards}~is a similar learning method that involves an auxiliary model.
However, in GAN the auxiliary model and the target model follow two \emph{opposite} goals,
while in CL the auxiliary model and the target model are models in the same track.
Moreover, in CL the auxiliary model is stable across the learning procedure,
while itself needs careful learning in GAN.

It's worth noting that
although our CL method bears certain level of resemblance with Noise Contrastive Estimation (NCE) \cite{gutmann2012noise}.
The motivation and the actual technical formulation of CL and NCE are essentially different.
For example,
in NCE the logistic function is a result of computing posterior probabilities,
while in CL it is explicitly introduced to saturate the influence of easy samples.

%benefit: suitable for all kinds of models
As CL requires only $p_m(c|I)$ and $p_n(c|I)$,
the choices of the target model and the reference model can
range from models based on LSTMs \cite{hochreiter1997long}~to models in other formats,
such as MRFs \cite{farhadi2010every}~and memory-networks \cite{park2017attend}.
On the other hand, although in CL,
 the reference model is usually fixed across the learning procedure,
one can replace the reference model with the latest target model periodically.
The reasons are
(1) $\nabla J(\vtheta) \ne \vzero$ when the target model and the reference model are identical,
(2) latest target model is usually stronger than the reference model,
(3) and a stronger reference model can provide stronger bounds and lead to a stronger target model.

\section{Experiment}
\label{sec:exprt}
\begin{table}
\centering
\small
\begin{tabular}{lccccccc}
\toprule
\multicolumn{8}{c}{COCO Online Testing Server C5} \\
\midrule
Method & B-1 & B-2 & B-3 & B-4 & METEOR & ROUGE\_L & CIDEr \\
\midrule
Google NIC \cite{vinyals2015show} & 0.713 & 0.542 & 0.407 & 0.309 & 0.254 & 0.530 & 0.943 \\
Hard-Attention\cite{xu2015show} & 0.705 & 0.528 & 0.383 & 0.277 & 0.241 & 0.516 & 0.865 \\
AdaptiveAttention \cite{lu2016knowing} & 0.735 & 0.569 & 0.429 & 0.323 & 0.258 & 0.541 & 1.001 \\ 
AdpativeAttention + CL (Ours) & 0.742 & 0.577 & 0.436 & 0.326 & \textbf{0.260} & 0.544 & 1.010 \\
PG-BCMR \cite{liu2016optimization} & \textbf{0.754} & \textbf{0.591} & \textbf{0.445} & \textbf{0.332} & 0.257 & \textbf{0.550} & \textbf{1.013} \\
\midrule
ATT-FCN$^\dagger$ \cite{you2016image} & 0.731 & 0.565 & 0.424 & 0.316 & 0.250 & 0.535 & 0.943 \\
MSM$^\dagger$ \cite{yao2016boosting} & 0.739 & 0.575 & 0.436 & 0.330 & 0.256 & 0.542 & 0.984 \\
AdaptiveAttention$^\dagger$ \cite{lu2016knowing} & 0.746 & 0.582 & 0.443 & 0.335 & 0.264 & 0.550 & 1.037 \\
Att2in$^\dagger$ \cite{rennie2016self} & - & - & - & 0.344 & 0.268 & 0.559 & 1.123 \\
\midrule
\midrule
\multicolumn{8}{c}{COCO Online Testing Server C40} \\
\midrule
Method & B-1 & B-2 & B-3 & B-4 & METEOR & ROUGE\_L & CIDEr \\
\midrule
Google NIC \cite{vinyals2015show} & 0.895 & 0.802 & 0.694 & 0.587 & 0.346 & 0.682 & 0.946 \\
Hard-Attention \cite{xu2015show} & 0.881 & 0.779 & 0.658 & 0.537 & 0.322 & 0.654 & 0.893 \\
AdaptiveAttention \cite{lu2016knowing} & 0.906 & 0.823 & 0.717 & 0.607 & 0.347 & 0.689 & 1.004 \\ 
AdaptiveAttention + CL (Ours) & \textbf{0.910} & \textbf{0.831} & \textbf{0.728} & \textbf{0.617} & \textbf{0.350} & \textbf{0.695} & \textbf{1.029} \\
PG-BCMR \cite{liu2016optimization} & - & - & - & - & - & - & - \\
\midrule
ATT-FCN$^\dagger$ \cite{you2016image} & 0.900 & 0.815 & 0.709 & 0.599 & 0.335 & 0.682 & 0.958 \\
MSM$^\dagger$ \cite{yao2016boosting} & 0.919 & 0.842 & 0.740 & 0.632 & 0.350 & 0.700 & 1.003 \\
AdaptiveAttention$^\dagger$ \cite{lu2016knowing} & 0.918 & 0.842 & 0.740 & 0.633 & 0.359 & 0.706 & 1.051 \\
Att2in$^\dagger$ \cite{rennie2016self} & - & - & - & - & - & - & - \\
\bottomrule
\end{tabular}
\vspace{1mm}
\caption{\small This table lists published results of state-of-the-art image captioning models on the online COCO testing server.
$\dagger$ indicates ensemble model. 
"-" indicates not reported.
In this table, CL improves the base model (AdaptiveAttention \cite{lu2016knowing}) to gain the best results among all single models on C40.}
\label{tab:cocoonline}
\vspace{-5mm}
\end{table}

\vspace{-5pt}
\subsection{Datasets}
We use two large scale datasets to test our contrastive learning method.
The first dataset is MSCOCO \cite{lin2014microsoft},
which contains $122,585$ images for training and validation.
Each image in MSCOCO has $5$ human annotated captions.
Following splits in \cite{lu2016knowing},
we reserved $2,000$ images for validation.
A more challenging dataset, InstaPIC-1.1M \cite{park2017attend}, 
is used as the second dataset,
which contains $648,761$ images for training, and $5,000$ images for testing.
The images and their ground-truth captions are acquired from Instagram,
where people post images with related descriptions.
Each image in InstaPIC-1.1M is paired with $1$ caption.
This dataset is challenging, as its captions are natural posts with varying formats.
In practice, we reserved $2,000$ images from the training set for validation.

On both datasets, non-alphabet characters except emojis are removed, 
and alphabet characters are converted to lowercases.
Words and emojis that appeared less than $5$ times are replaced with \emph{UNK}.
And all captions are truncated to have at most $18$ words and emojis.
As a result, we obtained a vocabulary of size $9,567$ on MSCOCO, and a vocabulary of size $22,886$ on InstaPIC-1.1M.

\vspace{-5pt}
\subsection{Settings}

To study the generalization ability of proposed CL method,
we tested it on two different image captioning models,
namely \textbf{Neuraltalk2} \cite{karpathy2015deep} and \textbf{AdaptiveAttention} \cite{lu2016knowing}. 
Both models are based on \emph{encoder-and-decoder} \cite{vinyals2015show},
where no attention mechanism is used in the former,
and an adaptive attention component is used in the latter.

For both models, we have pretrained them by MLE,
and use the pretrain checkpoints as initializations. 
In all experiments except for the experiment on model choices, 
we choose the same model and use the same initialization 
for target model and reference model. 
In all our experiments, 
we fixed the learning rate to be $1e$-$6$ for all components,
and used Adam optimizer. 
Seven evaluation metrics have been selected to compare the performances of different models,
including Bleu-1,2,3,4 \cite{papineni2002bleu}, Meteor \cite{lavie2014meteor}, Rouge \cite{lin2004rouge} and Cider \cite{vedantam2015cider}.
All experiments for ablation studies are conducted on the validation set of MSCOCO.

\begin{figure}
\centering
\includegraphics[width=\textwidth]{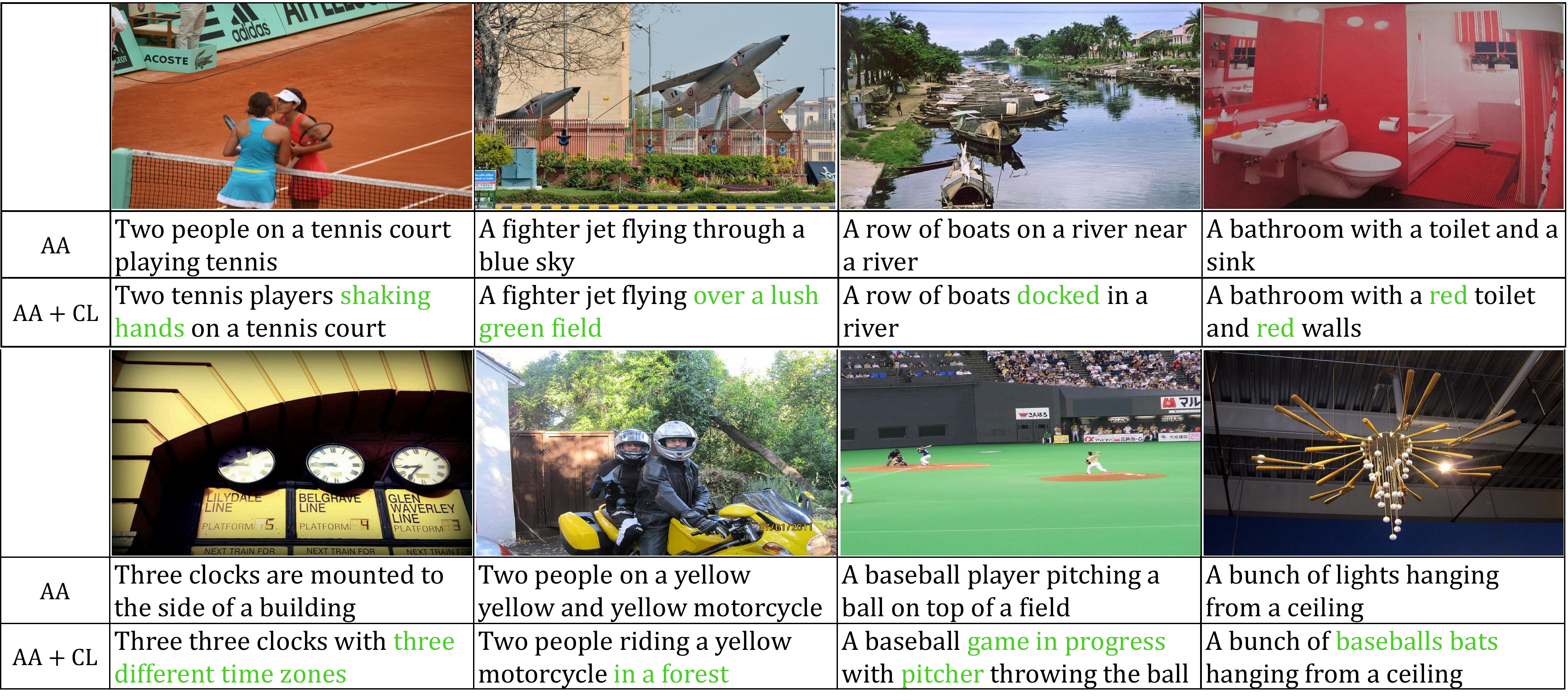}
\caption{\small This figure illustrates several images with captions generated by different models,
where \emph{AA} represents AdaptiveAttention \cite{lu2016knowing} learned by MLE, 
and \emph{AA + CL} represents the same model learned by CL.
Compared to \emph{AA}, \emph{AA + CL} generated more distinctive captions for these images.}
\label{fig:compare}
\end{figure}

\subsection{Results}
\begin{table}
\small
\centering
\begin{tabular}{lccccccc}
\toprule
Method & B-1 & B-2 & B-3 & B-4 & METEOR & ROUGE\_L & CIDEr \\
\midrule
Google NIC \cite{vinyals2015show} & 0.055 & 0.019 & 0.007 & 0.003 & 0.038 & 0.081 & 0.004 \\
Hard-Attention \cite{xu2015show} & 0.106 & 0.015 & 0.000 & 0.000 & 0.026 & 0.140 & 0.049 \\
CSMN \cite{park2017attend} & \textbf{0.079} & \textbf{0.032} & \textbf{0.015} & \textbf{0.008} & \textbf{0.037} & \textbf{0.120} & 0.133 \\
AdaptiveAttention \cite{lu2016knowing} & 0.065 & 0.026 & 0.011 & 0.005 & 0.029 & 0.093 & 0.126 \\
AdaptiveAttention + CL (Ours) & 0.072 & 0.028 & 0.013 & 0.006 & 0.032 & 0.101 & \textbf{0.144} \\
\bottomrule
\end{tabular}
\vspace{1mm}
\caption{\small This table lists results of different models on the test split of InstaPIC-1.1M \cite{park2017attend},
where CL improves the base model (AdaptiveAttention \cite{lu2016knowing}) by significant margins,
achieving the best result on Cider.}
\vspace{-2mm}
\end{table}

\vspace{-5pt}
\paragraph{Overall Results} We compared our best model (\emph{AdaptiveAttention} \cite{lu2016knowing} learned by CL) with state-of-the-art models on two datasets.
On MSCOCO, we submitted the results to the online COCO testing server.
%\footnote{\url{https://competitions.codalab.org/competitions/3221}}
The results along with other published results are listed in Table \ref{tab:cocoonline}.
Compared to MLE-learned \emph{AdaptiveAttention}, 
CL improves the performace of it by significant margins across all metrics.
While most of state-of-the-art results are achieved by ensembling multiple models,
our improved \emph{AdaptiveAttention} gains competitive results as a \emph{single} model.
Specifically, on Cider, CL improves \emph{AdaptiveAttention} from $1.003$ to $1.029$,
which is the best single-model result on C40 among all published ones.
In terms of Cider,
if we use MLE,
we need to combine $5$ models to get $4.5\%$ boost on C40 for \emph{AdaptiveAttention}.
Using CL, we improve the performance by $2.5\%$ with just a single model.
On InstaPIC-1.1M, CL improves the performance of \emph{AdaptiveAttention} by $14\%$ in terms of Cider,
which is the state-of-the-art.
Some qualitative results are shown in Figure \ref{fig:compare}.
It's worth noting that the proposed learning method can be used with stronger base models to obtain better results
without any modification.
\begin{table}
\centering
\small
\begin{tabular}{lccccccc}
\toprule
Method & B-1 & B-2 & B-3 & B-4 & METEOR & ROUGE\_L & CIDEr \\
\midrule
AdaptiveAttention \cite{lu2016knowing} (Base) & 0.733 & 0.572 & 0.433 & 0.327 & 0.260 & 0.540 & 1.042 \\
Base + IL \cite{vedantam2017context} & 0.706 & 0.544 & 0.408 & 0.307 & 0.253 & 0.530 & 1.004 \\
Base + GAN \cite{dai2017towards} & 0.629 & 0.437 & 0.290 & 0.190 & 0.212 & 0.458 & 0.700 \\
Base + CL(P) & 0.735 & 0.573 & 0.437 & 0.334 & 0.262 & 0.545 & 1.059 \\
Base + CL(N) & 0.539 & 0.411 & 0.299 & 0.212 & 0.246 & 0.479 & 0.603 \\
Base + CL(Full) & \textbf{0.755} & \textbf{0.598} & \textbf{0.460} & \textbf{0.353} & \textbf{0.271} & \textbf{0.559} & \textbf{1.142} \\
\bottomrule
\end{tabular}
\vspace{1mm}
\caption{\small This table lists results of a model learned by different methods.
The best result is obtained by the one learned with full CL, containing both the positive constraint and negative constraint.}
\label{tab:methods}
\vspace{-5mm}
\end{table}

\vspace{-5pt}
\paragraph{Compare Learning Methods}
Using \emph{AdaptiveAttention} learned by MLE as base model and initialization, 
we compared our CL with similar learning methods,
including \textbf{CL(P)} and \textbf{CL(N)} that respectively contains only the positive constraint and the negative constraint in CL.
We also compared with
\textbf{IL} \cite{vedantam2017context}, and \textbf{GAN} \cite{dai2017towards}.
The results on MSCOCO are listed in Table \ref{tab:methods},
where (1) among IL, CL and GAN, CL improves performance of the base model,
while both IL and GAN decrease the results. 
This indicates the trade-off between learning distinctiveness and maintaining overall performance 
is not well settled in IL and GAN.
(2) comparing models learned by CL(P), CL(N) and CL,
we found using the positive constraint or the negative constraint alone is not sufficient,
as only one source of guidance is provided.
While CL(P) gives the base model lower improvement than full CL, 
CL(N) downgrades the base model, 
indicating overfits on distinctiveness.
Combining CL(P) and CL(N), 
CL is able to encourage distinctiveness while also emphasizing on overall performance,
resulting in largest improvements on all metrics.

\begin{table}
\small
\centering
\begin{tabular}{ccccccccc}
\toprule
    Target Model & Reference Model & B-1 & B-2 & B-3 & B-4 & METEOR & ROUGE\_L & CIDEr \\
\midrule
    NT & - & 0.697 & 0.525 & 0.389 & 0.291 & 0.238 & 0.516 & 0.882\\
    NT & NT & 0.708 & 0.536 & 0.399 & 0.300 & 0.242 & 0.524 & 0.905 \\
    NT & AA & \textbf{0.716} & \textbf{0.547} & \textbf{0.411} & \textbf{0.311} & \textbf{0.249} & \textbf{0.533} & \textbf{0.956} \\
\midrule
    AA & - & 0.733 & 0.572 & 0.433 & 0.327 & 0.260 & 0.540 & 1.042 \\
    AA & AA & \textbf{0.755} & \textbf{0.598} & \textbf{0.460} & \textbf{0.353} & \textbf{0.271} & \textbf{0.559} & \textbf{1.142} \\
\bottomrule
\end{tabular}
\vspace{1mm}
\caption{\small This table lists results of different model choices on MSCOCO.
In this table, NT represents Neuraltalk2 \cite{karpathy2015deep}, and AA represents AdaptiveAttention \cite{lu2016knowing}.
"-" indicates the target model is learned using MLE.}
\label{tab:modelchoice}
\vspace{-2mm} 
\end{table}

\begin{table}
\small
\centering
\begin{tabular}{cccccccc}
\toprule
Run & B-1 & B-2 & B-3 & B-4 & METEOR & ROUGE\_L & CIDEr \\
\midrule
0 & 0.733 & 0.572 & 0.433 & 0.327 & 0.260 & 0.540 & 1.042 \\
1 & 0.755 & 0.598 & 0.460 & 0.353 & 0.271 & 0.559 & 1.142 \\
2 & 0.756 & 0.598 & 0.460 & 0.353 & 0.272 & 0.559 & 1.142 \\
\bottomrule
\end{tabular}
\vspace{1mm}
\caption{\small This table lists results of periodical replacement of the reference in CL.
In run 0, the model is learned by MLE,
which are used as both the target and the reference in run 1.
In run 2, the reference is replaced with the best target in run 1.}
\label{tab:periodic}
\vspace{-5mm}
\end{table}

\vspace{-5pt}
\paragraph{Compare Model Choices} To study the generalization ability of CL,
\emph{AdaptiveAttention} and \emph{Neuraltalk2} are respectively chosen 
as both the target and the reference in CL.
In addition, \emph{AdaptiveAttention} learned by MLE,
as a better model, is chosen to be the reference,
for \emph{Neuraltalk2}.
The results are listed in Table \ref{tab:modelchoice},
where compared to models learned by MLE,
both \emph{AdaptiveAttention} and \emph{Neuraltalk2} are improved after learning using CL.
For example, on Cider, \emph{AdaptiveAttention} improves from $1.042$ to $1.142$,
and \emph{Neuraltalk2} improves from $0.882$ to $0.905$.
Moreover, by using a stronger model, \emph{AdaptiveAttention}, as the reference,
\emph{Neuraltalk2} improves further from $0.905$ to $0.956$,
which indicates stronger references empirically provide tighter bounds on both the positive constraint and the negative constraint.

\vspace{-5pt}
\paragraph{Reference Replacement} As discussed in sec \ref{sec:discuss},
one can periodically replace the reference with latest best target model,
to further improve the performance.
In our study, using \emph{AdaptiveAttention} learned by MLE as a start,
each run we fix the reference model util the target saturates its performance on the validation set,
then we replace the reference with latest best target model 
and rerun the learning.
As listed in Table \ref{tab:periodic},
in second run, the relative improvements of the target model is incremental,
compared to its improvement in the first run.
Therefore, when learning a model using CL,
with a sufficiently strong reference,
the improvement is usually saturated in the first run,
and there is no need, in terms of overall performance,
to replace the reference multiple times.

\section{Conclusion}
\label{sec:concls}

In this paper, we propose Contrastive Learning,
a new learning method for image captioning.
By employing a state-of-the-art model as a reference,
the proposed method is able to maintain the optimality of the target model,
while encouraging it to learn from distinctiveness,
which is an important property of high quality captions.
On two challenging datasets, namely MSCOCO and InstaPIC-1.1M,
the proposed method improves the target model by significant margins,
and gains state-of-the-art results across multiple metrics.
On comparative studies,
the proposed method extends well to models with different structures,
which clearly shows its generalization ability.

\paragraph{Acknowledgment}
This work is partially supported by
the Big Data Collaboration Research grant from SenseTime Group (CUHK Agreement No.TS1610626), 
the General Research Fund (GRF) of Hong Kong (No.14236516)
and the Early Career Scheme (ECS) of Hong Kong (No.24204215).

{
\bibliographystyle{plain}
\bibliography{cl}

\begin{thebibliography}{10}

\bibitem{dai2017towards}
Bo~Dai, Sanja Fidler, Raquel Urtasun, and Dahua Lin.
\newblock Towards diverse and natural image descriptions via a conditional gan.
\newblock In {\em Proceedings of the IEEE International Conference on Computer
  Vision}, 2017.

\bibitem{devlin2015exploring}
Jacob Devlin, Saurabh Gupta, Ross Girshick, Margaret Mitchell, and C~Lawrence
  Zitnick.
\newblock Exploring nearest neighbor approaches for image captioning.
\newblock {\em arXiv preprint arXiv:1505.04467}, 2015.

\bibitem{fang2015captions}
Hao Fang, Saurabh Gupta, Forrest Iandola, Rupesh~K Srivastava, Li~Deng, Piotr
  Doll{\'a}r, Jianfeng Gao, Xiaodong He, Margaret Mitchell, John~C Platt,
  et~al.
\newblock From captions to visual concepts and back.
\newblock In {\em Proceedings of the IEEE Conference on Computer Vision and
  Pattern Recognition}, pages 1473--1482, 2015.

\bibitem{farhadi2010every}
Ali Farhadi, Mohsen Hejrati, Mohammad~Amin Sadeghi, Peter Young, Cyrus
  Rashtchian, Julia Hockenmaier, and David Forsyth.
\newblock Every picture tells a story: Generating sentences from images.
\newblock In {\em European conference on computer vision}, pages 15--29.
  Springer, 2010.

\bibitem{gutmann2012noise}
Michael~U Gutmann and Aapo Hyv{\"a}rinen.
\newblock Noise-contrastive estimation of unnormalized statistical models, with
  applications to natural image statistics.
\newblock {\em Journal of Machine Learning Research}, 13(Feb):307--361, 2012.

\bibitem{hochreiter1997long}
Sepp Hochreiter and J{\"u}rgen Schmidhuber.
\newblock Long short-term memory.
\newblock {\em Neural computation}, 9(8):1735--1780, 1997.

\bibitem{jas2015image}
Mainak Jas and Devi Parikh.
\newblock Image specificity.
\newblock In {\em Proceedings of the IEEE Conference on Computer Vision and
  Pattern Recognition}, pages 2727--2736, 2015.

\bibitem{karpathy2015deep}
Andrej Karpathy and Li~Fei-Fei.
\newblock Deep visual-semantic alignments for generating image descriptions.
\newblock In {\em Proceedings of the IEEE Conference on Computer Vision and
  Pattern Recognition}, pages 3128--3137, 2015.

\bibitem{kulkarni2013babytalk}
Girish Kulkarni, Visruth Premraj, Vicente Ordonez, Sagnik Dhar, Siming Li,
  Yejin Choi, Alexander~C Berg, and Tamara~L Berg.
\newblock Babytalk: Understanding and generating simple image descriptions.
\newblock {\em IEEE Transactions on Pattern Analysis and Machine Intelligence},
  35(12):2891--2903, 2013.

\bibitem{kuznetsova2014treetalk}
Polina Kuznetsova, Vicente Ordonez, Tamara~L Berg, and Yejin Choi.
\newblock Treetalk: Composition and compression of trees for image
  descriptions.
\newblock {\em TACL}, 2(10):351--362, 2014.

\bibitem{lavie2014meteor}
Michael Denkowski~Alon Lavie.
\newblock Meteor universal: Language specific translation evaluation for any
  target language.
\newblock {\em ACL 2014}, page 376, 2014.

\bibitem{lin2004rouge}
Chin-Yew Lin.
\newblock Rouge: A package for automatic evaluation of summaries.
\newblock In {\em Text summarization branches out: Proceedings of the ACL-04
  workshop}, volume~8. Barcelona, Spain, 2004.

\bibitem{lin2014microsoft}
Tsung-Yi Lin, Michael Maire, Serge Belongie, James Hays, Pietro Perona, Deva
  Ramanan, Piotr Doll{\'a}r, and C~Lawrence Zitnick.
\newblock Microsoft coco: Common objects in context.
\newblock In {\em European Conference on Computer Vision}, pages 740--755.
  Springer, 2014.

\bibitem{liu2016optimization}
Siqi Liu, Zhenhai Zhu, Ning Ye, Sergio Guadarrama, and Kevin Murphy.
\newblock Optimization of image description metrics using policy gradient
  methods.
\newblock {\em arXiv preprint arXiv:1612.00370}, 2016.

\bibitem{lu2016knowing}
Jiasen Lu, Caiming Xiong, Devi Parikh, and Richard Socher.
\newblock Knowing when to look: Adaptive attention via a visual sentinel for
  image captioning.
\newblock {\em arXiv preprint arXiv:1612.01887}, 2016.

\bibitem{mao2016generation}
Junhua Mao, Jonathan Huang, Alexander Toshev, Oana Camburu, Alan~L Yuille, and
  Kevin Murphy.
\newblock Generation and comprehension of unambiguous object descriptions.
\newblock In {\em Proceedings of the IEEE Conference on Computer Vision and
  Pattern Recognition}, pages 11--20, 2016.

\bibitem{papineni2002bleu}
Kishore Papineni, Salim Roukos, Todd Ward, and Wei-Jing Zhu.
\newblock Bleu: a method for automatic evaluation of machine translation.
\newblock In {\em Proceedings of the 40th annual meeting on association for
  computational linguistics}, pages 311--318. Association for Computational
  Linguistics, 2002.

\bibitem{park2017attend}
Cesc~Chunseong Park, Byeongchang Kim, and Gunhee Kim.
\newblock Attend to you: Personalized image captioning with context sequence
  memory networks.
\newblock In {\em CVPR}, 2017.

\bibitem{rennie2016self}
Steven~J Rennie, Etienne Marcheret, Youssef Mroueh, Jarret Ross, and Vaibhava
  Goel.
\newblock Self-critical sequence training for image captioning.
\newblock {\em arXiv preprint arXiv:1612.00563}, 2016.

\bibitem{simonyan2014very}
Karen Simonyan and Andrew Zisserman.
\newblock Very deep convolutional networks for large-scale image recognition.
\newblock {\em arXiv preprint arXiv:1409.1556}, 2014.

\bibitem{vedantam2017context}
Ramakrishna Vedantam, Samy Bengio, Kevin Murphy, Devi Parikh, and Gal Chechik.
\newblock Context-aware captions from context-agnostic supervision.
\newblock {\em arXiv preprint arXiv:1701.02870}, 2017.

\bibitem{vedantam2015cider}
Ramakrishna Vedantam, C~Lawrence~Zitnick, and Devi Parikh.
\newblock Cider: Consensus-based image description evaluation.
\newblock In {\em Proceedings of the IEEE Conference on Computer Vision and
  Pattern Recognition}, pages 4566--4575, 2015.

\bibitem{vinyals2015show}
Oriol Vinyals, Alexander Toshev, Samy Bengio, and Dumitru Erhan.
\newblock Show and tell: A neural image caption generator.
\newblock In {\em Proceedings of the IEEE Conference on Computer Vision and
  Pattern Recognition}, pages 3156--3164, 2015.

\bibitem{xu2015show}
Kelvin Xu, Jimmy Ba, Ryan Kiros, Kyunghyun Cho, Aaron~C Courville, Ruslan
  Salakhutdinov, Richard~S Zemel, and Yoshua Bengio.
\newblock Show, attend and tell: Neural image caption generation with visual
  attention.
\newblock In {\em ICML}, volume~14, pages 77--81, 2015.

\bibitem{yao2016boosting}
Ting Yao, Yingwei Pan, Yehao Li, Zhaofan Qiu, and Tao Mei.
\newblock Boosting image captioning with attributes.
\newblock {\em arXiv preprint arXiv:1611.01646}, 2016.

\bibitem{you2016image}
Quanzeng You, Hailin Jin, Zhaowen Wang, Chen Fang, and Jiebo Luo.
\newblock Image captioning with semantic attention.
\newblock In {\em Proceedings of the IEEE Conference on Computer Vision and
  Pattern Recognition}, pages 4651--4659, 2016.

\bibitem{zhou2016image}
Luowei Zhou, Chenliang Xu, Parker Koch, and Jason~J Corso.
\newblock Image caption generation with text-conditional semantic attention.
\newblock {\em arXiv preprint arXiv:1606.04621}, 2016.

\end{thebibliography}
}
\end{document}